\documentclass{article}

    \usepackage[nonatbib, preprint]{neurips_2019}

\usepackage{times}
\usepackage{epsfig}
\usepackage{graphicx}
\usepackage{amsmath}
\usepackage{amssymb}
\usepackage{arydshln}
\usepackage[flushleft]{threeparttable}

\usepackage[utf8]{inputenc} % allow utf-8 input
\usepackage[T1]{fontenc}    % use 8-bit T1 fonts
\usepackage{hyperref}       % hyperlinks
\usepackage{url}            % simple URL typesetting
\usepackage{booktabs}       % professional-quality tables
\usepackage{amsfonts}       % blackboard math symbols
\usepackage{nicefrac}       % compact symbols for 1/2, etc.
\usepackage{microtype}      % microtypography

\DeclareMathOperator*{\argmax}{argmax} % thin space, limits underneath in displays
\def\bO{{\mathbf O}}
\def\bI{{\mathbf I}}
\def\bW{{\mathbf W}}
\def\bF{{\mathbf F}}
\def\bv{{\mathbf v}}
\def\cL{{\mathcal L}}
\newcommand{\etal}{\textit{et al}.}

\graphicspath{ {./images/} }

\title{Automatic Pruning for Quantized Neural Networks}

\author{%
   Luis Guerra$^1$
   \And
   Bohan Zhuang$^2$
   \And
   Ian Reid$^2$
   \And
   Tom Drummond$^1$\\
   \authorend
   $^1$Monash University
   \and
   $^2$The University of Adelaide
}

\begin{document}

\maketitle

\begin{abstract}
Neural network quantization and pruning are two techniques commonly used to reduce the computational complexity and memory footprint of these models for deployment. However, most existing pruning strategies operate on full-precision and cannot be directly applied to discrete parameter distributions after quantization. In contrast, we study a combination of these two techniques to achieve further network compression. In particular, we propose an effective pruning strategy for selecting redundant low-precision filters. Furthermore, we leverage Bayesian optimization to efficiently determine the pruning ratio for each layer. We conduct extensive experiments on CIFAR-10 and ImageNet with various architectures and precisions. In particular, for ResNet-18 on ImageNet, we prune 26.12\% of the model size with Binarized Neural Network quantization, achieving a top-1 classification accuracy of 47.32\% in a model of 2.47 MB and 59.30\% with a 2-bit DoReFa-Net in 4.36 MB.
\end{abstract}

\section{Introduction}

In recent years, Deep Neural Networks (DNNs) have driven scientific research in multiple Artificial Intelligence (AI) fields (\textit{e.g.}, computer vision \cite{long2015fully, he2016deep, lin2017refinenet}, natural language processing \cite{sutskever2014sequence, vaswani2017attention},  and audio processing). However, DNNs capabilities still do not translate into end-user applications on resource constrained mobile and embedded devices. Therefore, reducing the gap between experimental and feasible real world implementations is an increasingly active topic of research. 

Traditional DNN architectures were design to improve accuracy in diverse tasks, hence, they are commonly highly overparameterized for the task at hand. For example, ResNet-50 for Imagenet contains 26 million parameters compared with ResNet-18, which has 11 million parameters at an expense of a small accuracy decrease. Additionally, full-precision parameters are not efficient or even required. Given the massive amount of Multiply-Accumulate (MAC) operations performed during dot-product calculations, parameter redundancy and floating-point operations represent the bottleneck towards low latency, memory and power efficient DNNs.

%In order to tackle this obstacles, a number of strategies have been proposed which can be grouped in five categories, namely, model compression, quantization, dynamic routing, Neural Architecture Search (NAS) and Automatic Machine Learning (AutoML). Model compression, emcompasses the subcategories of network pruning, matrix decomposition and weight-sharing. Quantized Neural Networks (QNN) and their special case, Binarized Neural Networks (BNN), aim at representing DNNs using limited discrete set of values. Dynamic routing considers the problem of taking decisions on which sections of the algorithm to compute at run-time based on the input data. Neural Architecture Search leverages on reinforcement learning to introduce the network design itself into the error-minimization process, it targets at discovering high-performing networks by stacking up DNNs building blocks (e.g. convolutional layers, residual connections, activation functions) in an optimal configuration. Not surpringly, an optimal architecture for latency might not be optimal for power consumption. AutoML algorithms goal is to isolate machine learning (ML) practitioners from the design cycle performing end-to-end optimization, ranging from data preprocessing to hyperparameter selection.

Quantization~\cite{courbariaux2015binaryconnect, hubara2016binarized, zhou2016dorefa, zhuang2018towards, zhang2018lq} and network pruning~\cite{li2016pruning, he2017channel, luo2017thinet} have been developed as effective solutions to tackle these obstacles.
With quantization, a low-precision model is deployed on devices, where operating in lower precision mode reduces computation as well as storage requirements. Moreover, with pruning, the redundant network modules are removed so that a compact network is achieved for inference. Although significant progress has been achieved by the previously described strategies, they are not mutually exclusive and little effort has been done towards combining their strengths. Tung and Mori \etal\cite{tung2018clip} previously considered this challenge, however they did it in a fine-grained fashion, this is, by pruning individual connections and quantizing. To the best of our knowledge Xu \etal\cite{xu2018main} investigated the problem of pruning structures for the first time, where they did it using a learning-based approach. Learning-based approaches commonly operate by partially \cite{louizos2017learning} or completely removing channels during training and progressively identifying the most discriminative ones. However, as observed in~\cite{liu2017learning}, removing channels affects the internal statistics of the network and consequently the real impact of removing such channel can only be indirectly measured in the loss. Thus, rule-based approaches are still actively investigated \cite{li2016pruning, zhuang2018discrimination}. In this paper we focus on the problem of pruning structures for quantized networks using a rule-based approach. Our strategy focuses on removing entire structures (\textit{i.e.} kernels and filters), making networks simpler to implement by avoiding sparse operations.
%\bohancomment{describe how we do and what's the advantage here} 
In particular, we propose a Quantized Neural Network (QNN) specific pruning approach based on the geometry of the QNN and the interactions between consecutive layers.

In our experiments, we prune redundant structures in already compact models without a high degradation in accuracy. Pruning QNNs can help improve generalization and yield extremely light-weight yet accurate DNNs with potential applications on real-time mobile and embedded devices.

%The Straight-Through Estimator (STE) has allowed to backpropagate error gradients through non-differentiable functions

\section{Related works}

\paragraph{ Network quantization.} 
%~\cite{bi-real} proposed a magnitude-aware gradient estimator allowing better convergence. ~\cite{self-binarizing} proposed to relax the sign function by replacing it with a hard hyperbolic tangent and progressively converging it into a sign function allowing to backpropagate gradients.  Recent quantization works~\cite{error-aware binarization, loss-error aware binarization} enforce the network to converge to a discrete weight distribution by integrating the quantization into the optimization step.
Network quantization aim at representing DNNs using a limited discrete set of values. Quantization can be categorized into fixed-point quantization and binary neural networks. Uniform approaches~\cite{zhou2016dorefa, zhuang2018towards} simply design quantizers with a constant quantization step. To reduce the quantization error, non-uniform strategies~\cite{zhang2018lq, jung2018joint, Cai_2017_CVPR} propose to jointly learn the quantizer and model parameters for better accuracy. Moreover, to relax the non-differentiable quantizer, which is core issue of quantization, some works propose to make the gradient-based optimization feasible by using gumble softmax~\cite{louizos2019relaxed} or learning with regularization~\cite{bai2019proxquant}. 
To further reduce the computational complexity, power consumption and memory storage, Binary Neural Networks (BNNs)~\cite{hubara2016binarized, rastegari2016xnor} propose to constrain weights and optionally activations to only two possible values (\textit{e.g.}, -1 or +1). When both weights and activations are binary, the multiply-accumulations can be replaced by the bitwise $\rm xnor(\cdot)$ and $\rm popcount(\cdot)$ operations. In order to solve the severe prediction accuracy degradation, some works have proposed multiple binarizations to approximate full-precision tensors~\cite{lin2017towards, li2017performance, guo2017network, tang2017train} or structures~\cite{zhuang2019structured}. Furthermore, Anderson and Berg \etal~\cite{anderson2017high} provide an explanation for the inference mechanism of BNNs generalizable to QNNs based on the \textit{angle preservation} and \textit{dot product preservation} properties of binary vectors with high dimensionality. Leroux~\etal \cite{leroux2019training} further improve the accuracy of a BNN by reducing the angles at each layer with a full-precision guidance network. Our pruning method is built on top of these concepts and looks towards reducing the angle between a real vector and its binary representation, consequently approximating dot products more accurately.

\paragraph{Network pruning.}
Network pruning is another dominant approach for compressing models which aims at removing redundant structures for accelerating run-time inference speed.
The pruning levels can be roughly categorized into filter pruning~\cite{li2016pruning, luo2017thinet, yu2018nisp}, channel pruning~\cite{he2017channel, liu2017learning, zhuang2018discrimination} and block pruning~\cite{wu2018blockdrop}. Most approaches are based on heuristic metrics which can be based on weight magnitude~\cite{he2017channel}, batch normalization scales~\cite{liu2017learning}, gradient magnitude~\cite{yu2018nisp, lee2019snip}.
To overcome handcraft hyperparameters, several works have proposed to leverage reinforcement learning to automatically prune channels~\cite{ashok2018n2n, he2018amc} or select blocks~\cite{wu2018blockdrop, wang2018skipnet}. For example, Wu~\etal~\cite{wu2018blockdrop} propose to learn a policy network in an associative reinforcement learning setting to dynamically choose parts of the network to execute.
Instead, we propose to use Bayesian optimization for channel pruning. Compared with reinforcement learning, Bayesian optimization methods~\cite{snoek2012practical} can make more efficient exploration
by estimating the density of good configurations based on a
probabilistic model.
More importantly, previous pruning literature focuses on compressing pretrained floating-point networks. 
In contrast, we investigate network pruning on quantized networks, whose weights and/or activations are discretized. We argue that quantized networks can be further pruned and propose effective pruning metrics accordingly to further save memory footprint and computational burden.
%~\cite{pruningchannelsforefficientconvnets}
%~\cite{clip-q net}

\section{Proposed method}
Given an $L$-layer QNN, let $\bF_k \in \mathbb{R}^{C\times h_{f}\times w_{f}}$ be the $k$-th convolutional filter in the $l$-th layer, where $C$, $h_{f}$ and $w_{f}$ denote the input channels, height and width of the filter respectively. Let $\bI \in \mathbb{R}^{C \times h_{in} \times w_{in}}$ and $\bO_k \in \mathbb{R}^{h_{out} \times w_{out}}$ be the input and output feature maps of the filter, where $h_{in}$, $w_{in}$, $h_{out}$, $w_{out}$ represent the height and width of the input and output feature maps, respectively. The output feature map of the $k$-th filter, $\bO_k$, is computed by convolving $\bI$ with $\bF_k$:
\begin{equation}
\bO_k=\sum_{c=1}^{C}\bI_{c,:,:}\ast \bF_{c,:,:}.
\label{eq:convolution}
\end{equation}

Training a QNN involves storing a shadow real-valued copy of the network which is updated during the optimization process by estimating the gradients w.r.t. the real weights. During inference, the real-valued network is quantized using a predetermined pointwise quantization function $\theta$ and the full-precision weights can be disposed. Given a pre-trained QNN, let the set of $K$ real-valued filters for the $l$-th layer be $\bW = \{\bF_1, ... , \bF_K \}$, our goal is to prune non-informative or even harmful filters aiming to save computational resources and reduce latency. In order to choose those filters, we propose a rule-based method to rank them by importance based on the interaction of each filter with the next layer (see Figure~\ref{fig:convolutional_interactions}). We will elaborate on how to measure the importance in Sec.~\ref{sec:metrics}. Then we will describe the pruning methods on individual kernels in Sec.~\ref{sec:kernels} and whole filters in Sec.~\ref{sec:filters}. Finally, we introduce how to automate the pruning process in Sec.~\ref{sec:ratio}.
%\bohancomment{don't understand what does the interactions mean here} \luiscomment{It will be more clear with the image}  

\subsection{Pruning metrics} \label{sec:metrics}
Common quantization sets, also knows as codebooks, do not contain a \textit{zero} value, forcing each real weight to land on an available quantization bin, implicitly injecting noise to the inference process. Intuitively, a real-value weight which lies near any quantization boundary will land on either quantization bin during stochastic quantization with approximately the same probability \cite{courbariaux2015binaryconnect}. Moreover, using deterministic quantization, these weights can easily be flipped into another bin during a gradient descent update step, harming the convergence of the optimization process.

Filters with multiple weights near the quantization boundaries will relate to a higher distance between the real-valued and the quantized versions. Additionally, a scaled version of a filter will have a zero angle w.r.t to the original one, thus proportional dot-product.
In particular, two metrics are considered to measure the distance between the real representation of a vector and the quantized one. 
A subset of filters is selected in the $l$-th layer given a metric and a threshold $th$:
\begin{equation}
\bW=\{ \bF_k|distance(\bF_k,\theta (\bF_k)) < th \}.
\label{eq:filter_subset}
\end{equation}

We re-arrange the $k$-th filter $\bF_k$ into a 1-dimensional vector $\bv$. The first ranking metric used is the angle, which can be computed from the cosine distance between the real-valued and the quantized vectors defined as
\begin{equation}
\phi = \frac{\bv \cdot \theta (\bv)}{\left \| \bv \right \| \cdot \left \| \theta (\bv) \right \|}.
\label{eq:angle_between_vectors}
\end{equation}

For the particular binary case it can efficiently computed as $\phi = \frac{\sum_i \left | v_i \right |}{\sqrt{\sum_i v_i^2} * \sqrt{n}}$.

Additionally, as second ranking metric, let the Euclidean distance between the two vectors be defined as
\begin{equation}
d = \sqrt{\sum_i (v_i - \theta (v_i ))^2}.
\label{eq:euclidean_distance}
\end{equation}

\subsection{Pruning individual kernels}  \label{sec:kernels}
\begin{figure}
    \centering
    \setlength{\tabcolsep}{-7.6pt}
    \resizebox{1.0\linewidth}{!}
    {\begin{tabular}{lcl}
    \includegraphics[width=5.3cm]{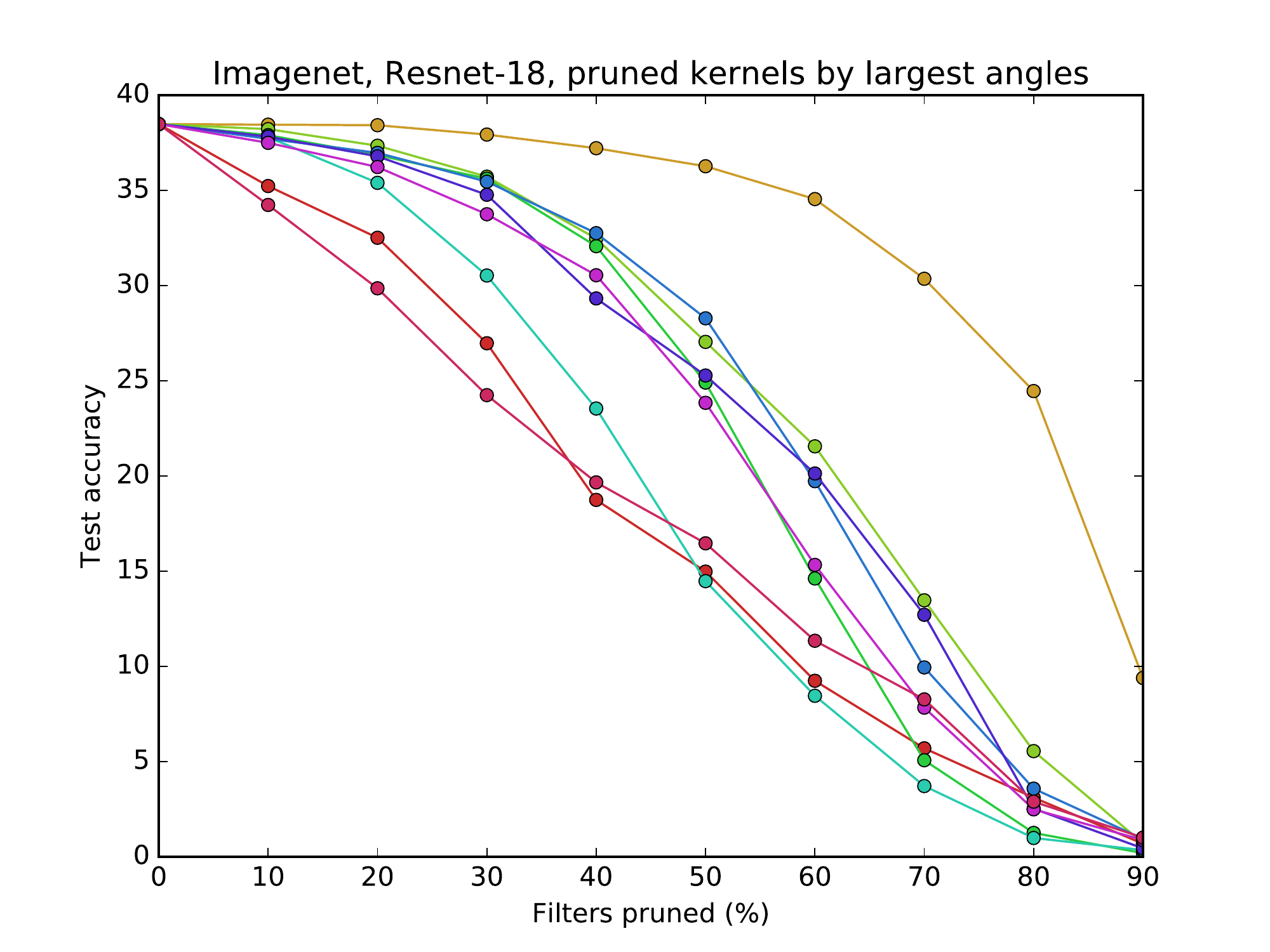}&
    \includegraphics[width=5.3cm]{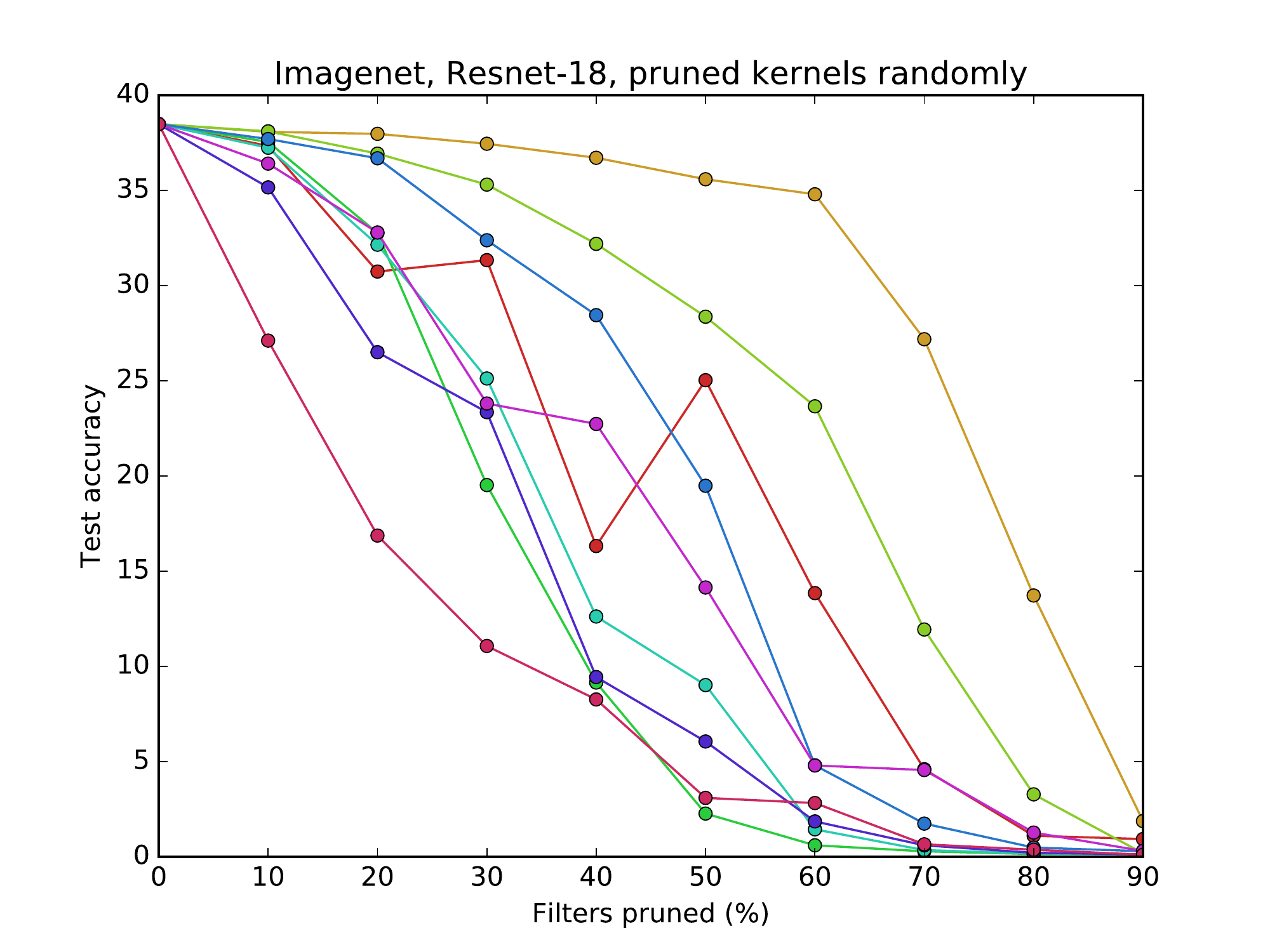}&
    \includegraphics[width=5.3cm]{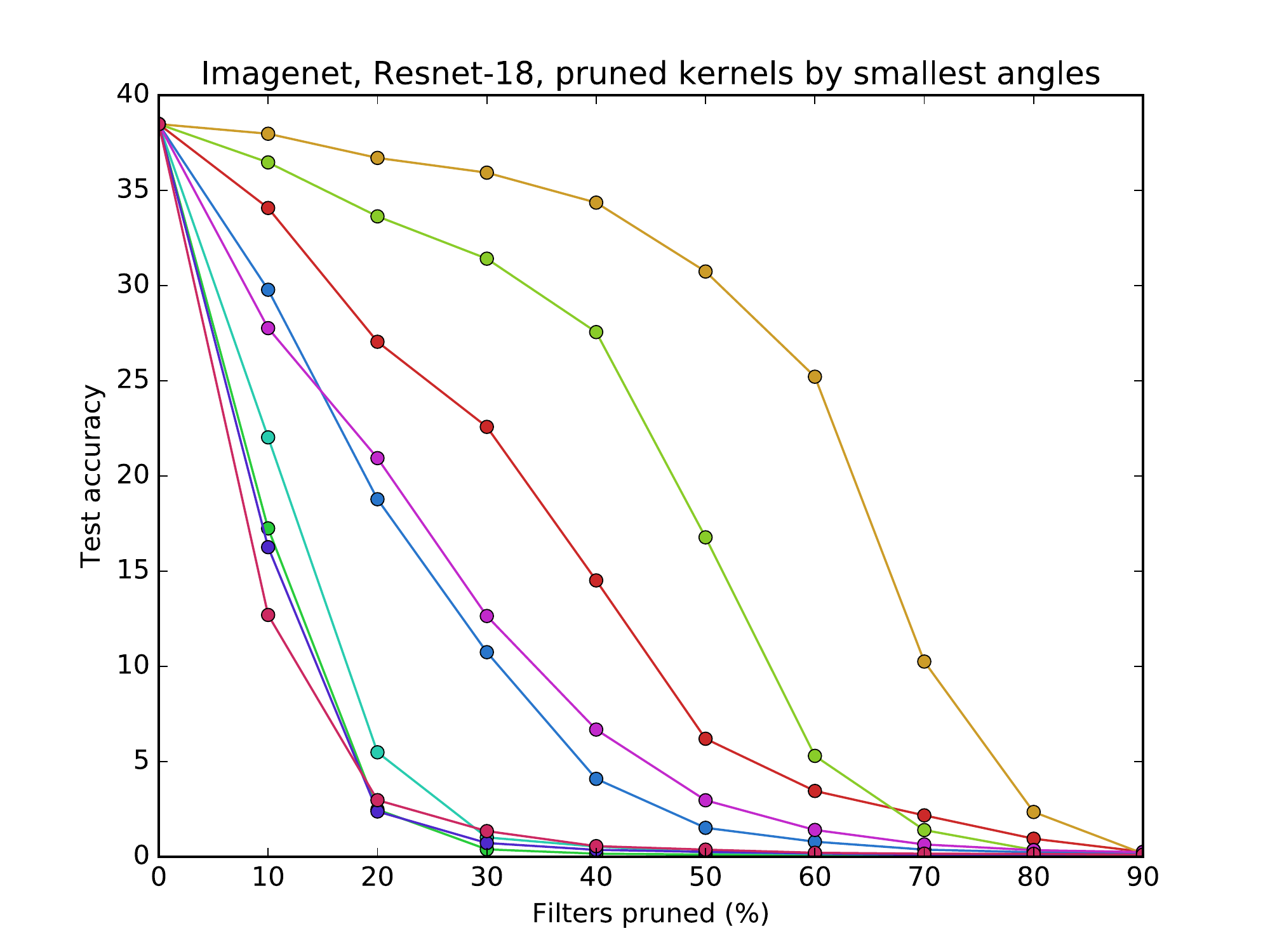}\\
    \includegraphics[width=5.3cm]{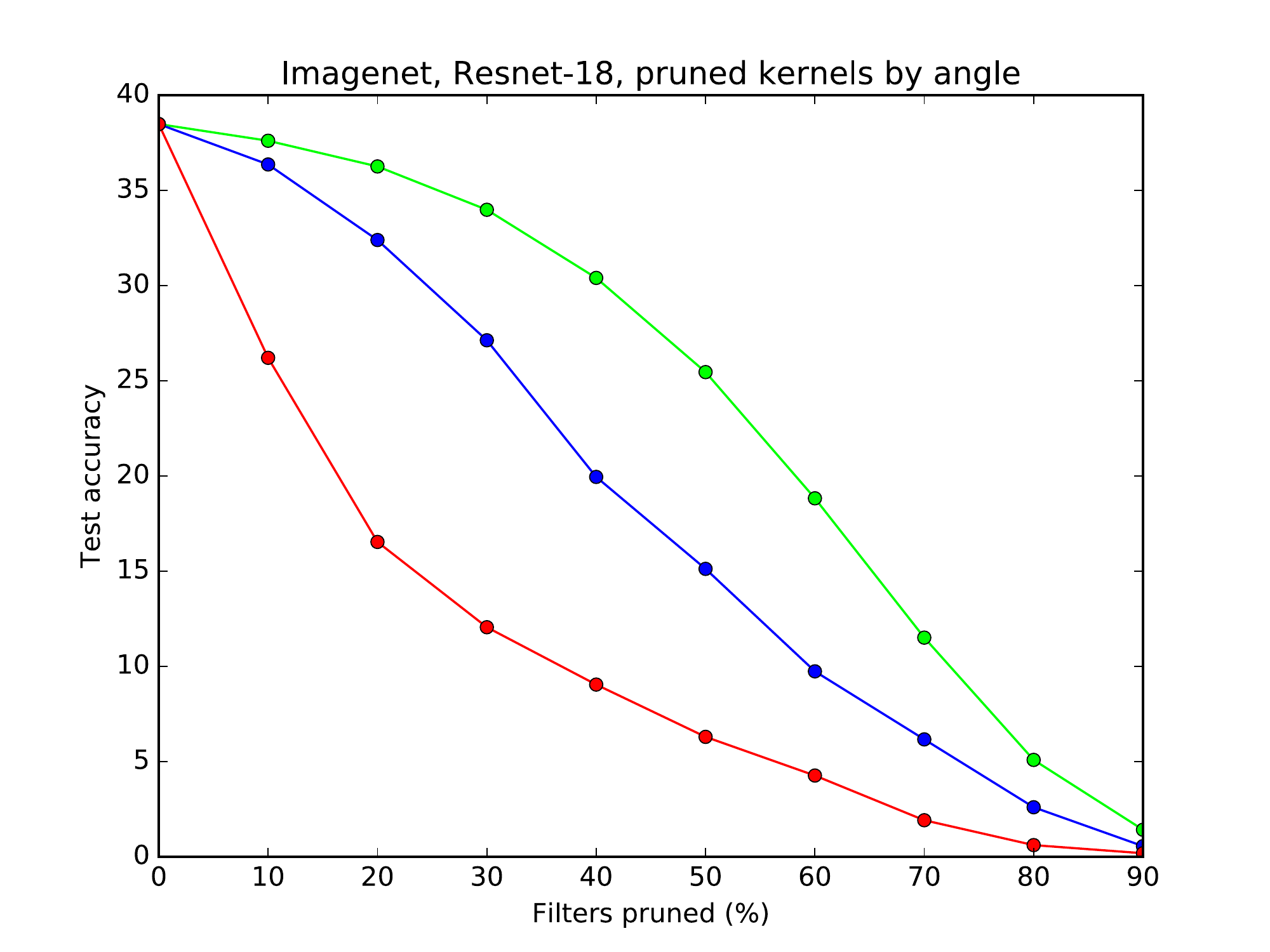}&
    \includegraphics[width=5.3cm]{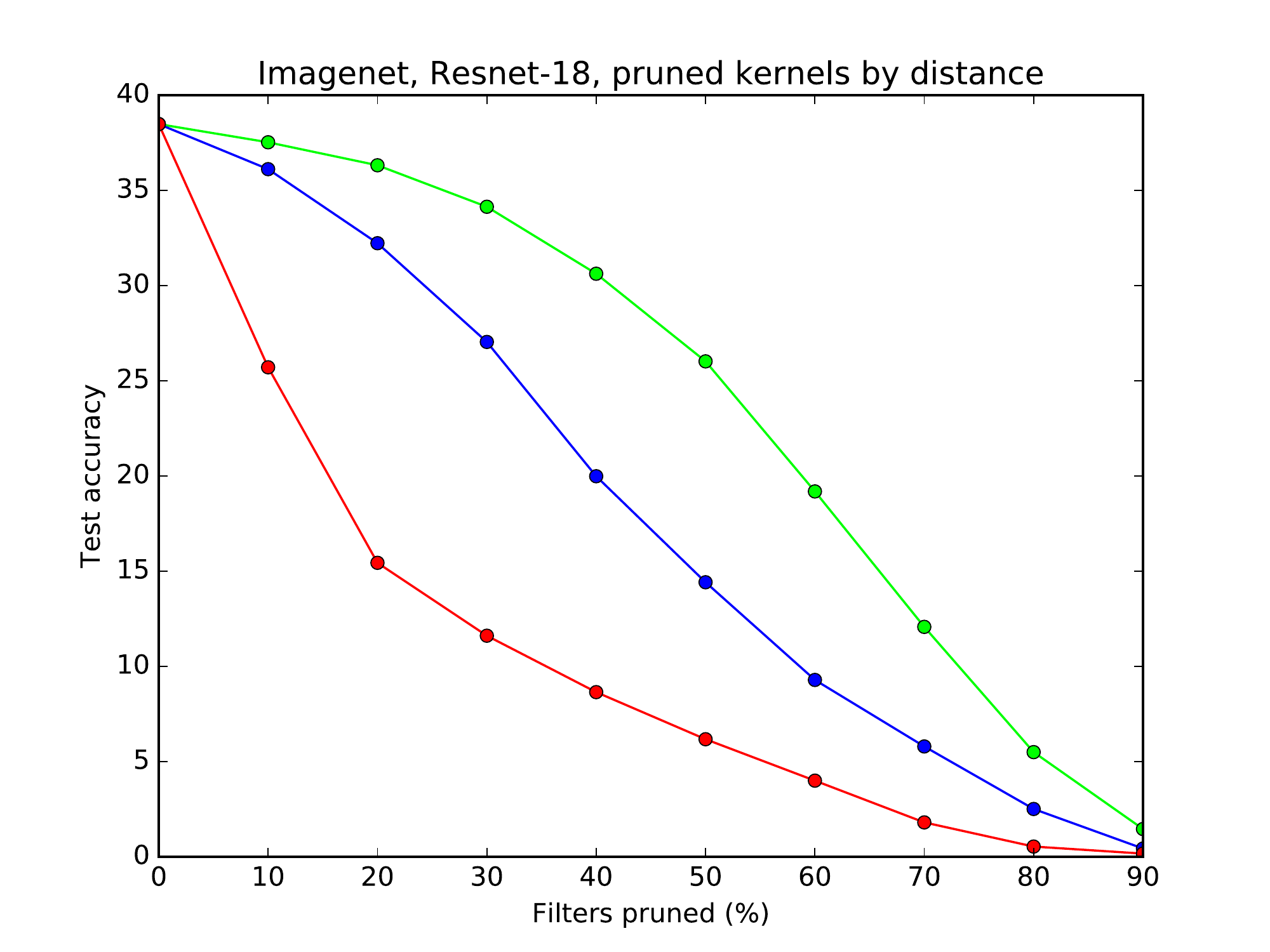}&
    \hbox{\hspace{1.6em}
    \includegraphics[width=2cm]{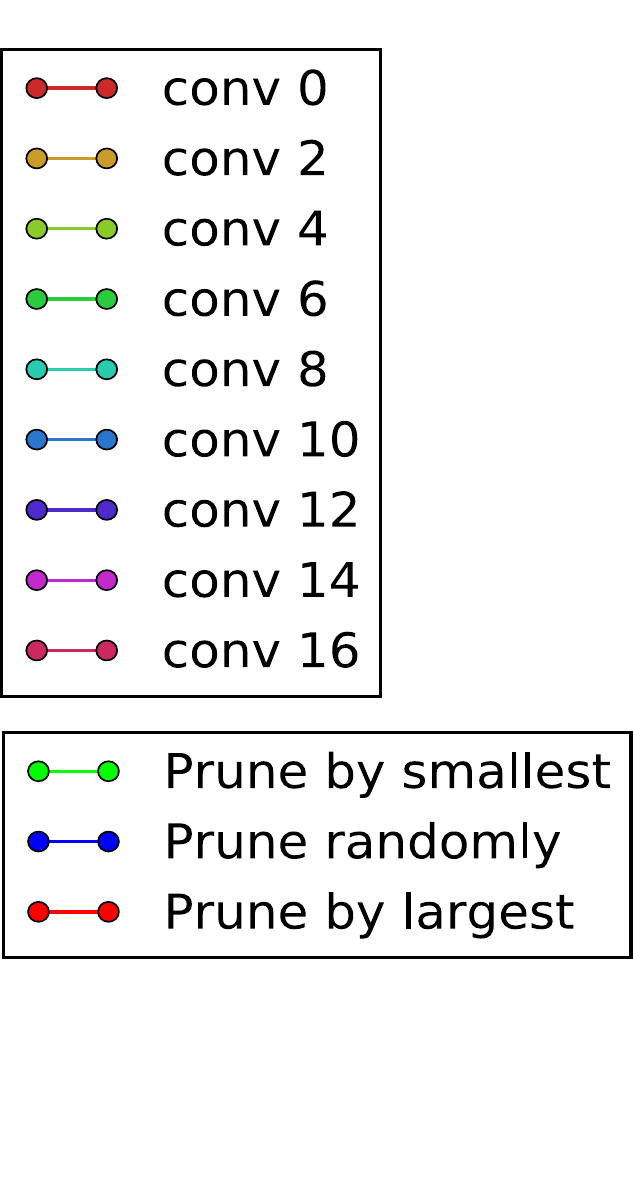}}
    \end{tabular}}
    \caption{Pruning individual kernels layer-wise in a BNN ResNet-18 trained on Imagenet. (a): Sorting by largest angle first and then pruning. (b): Pruning randomly. (c): Sorting by smallest angle first and then pruning. (d): Accuracy change during pruning averaged across all the layers in order to provide a cleaner visualization. It uses angle distance as metric. (e): Same as (d) but using euclidean distance as metric.}
    \label{fig:pruning_kernels}
\end{figure}

Every convolutional layer is comprised by $K$ filters with $C$ input channels as shown in Figure~\ref{fig:convolutional_interactions}, thus it contains $C\times K$ individual kernels.
As an initial experiment, we first rank and prune the individual kernels in the $l$-th layer computing the distances kernel-wise. In Figure~\ref{fig:pruning_kernels}, kernels were sorted by distance according to Eq.~(\ref{eq:angle_between_vectors}) and Eq.~(\ref{eq:euclidean_distance}) and pruned in descendent, random and ascending order, it can be observed that both metrics are useful for ranking the kernels and yield similar results.
Additionally, it can also be observed that some layers are more sensitive to pruning than others. Automatically sampling a near optimal pruning threshold will be addressed in Sec.~\ref{subsec:bayesian_optimization}.

\subsection{Pruning filters}  \label{sec:filters}
\begin{figure}
    \centering
    \resizebox{0.8\linewidth}{!}
    {\begin{tabular}{c}
    \includegraphics{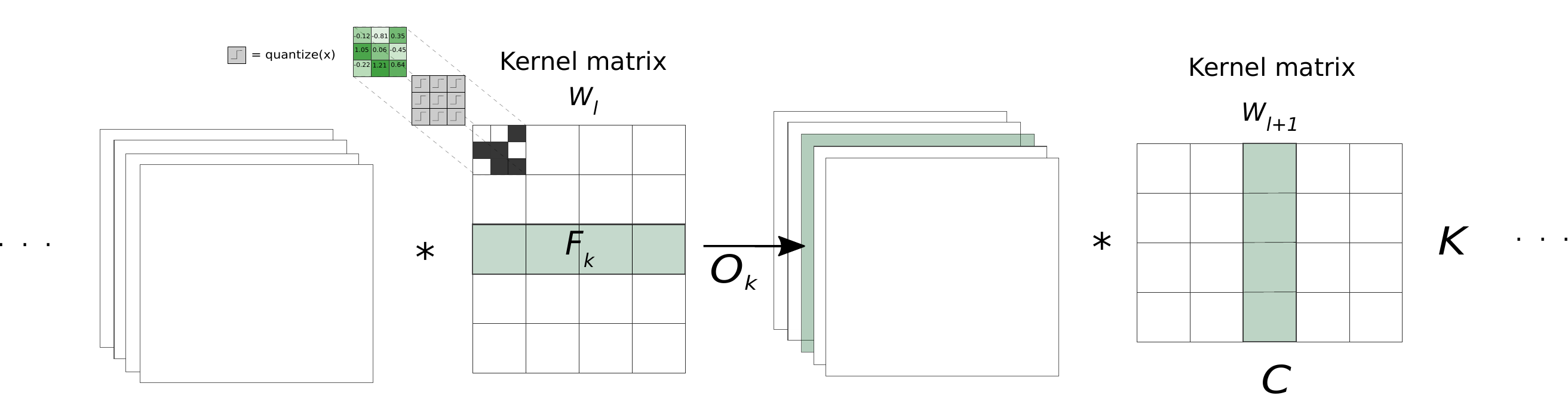}
    \end{tabular}}
    \caption{The $l$-th layer is conformed by $K\times C$ kernels. Kernel-wise distances can be averaged across the output channels in the $l$-th layer or across the input channels in the ($l+1$)-th layer.} %\bohancomment{cannot understand this figure and where's $l$-th layer? There should be $k$-th and $(k+1)$ layers, please refer to \url{https://arxiv.org/pdf/1608.08710.pdf}}}
    \label{fig:convolutional_interactions}
\end{figure}

\begin{figure}
    \centering
    \setlength{\tabcolsep}{0pt}
    \resizebox{0.9\linewidth}{!}    
    {\begin{tabular}{lcc}
    \includegraphics{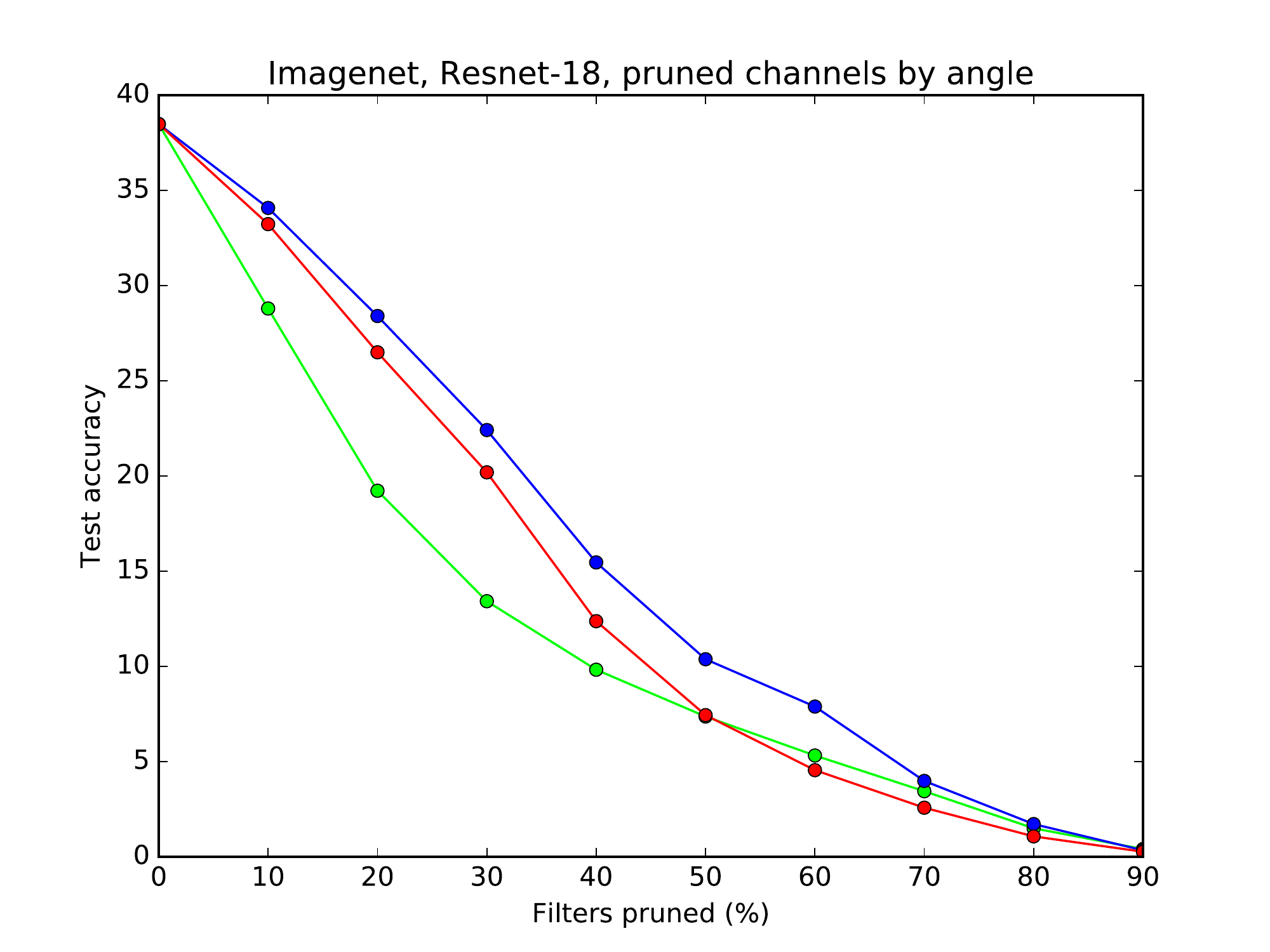}&
    \includegraphics{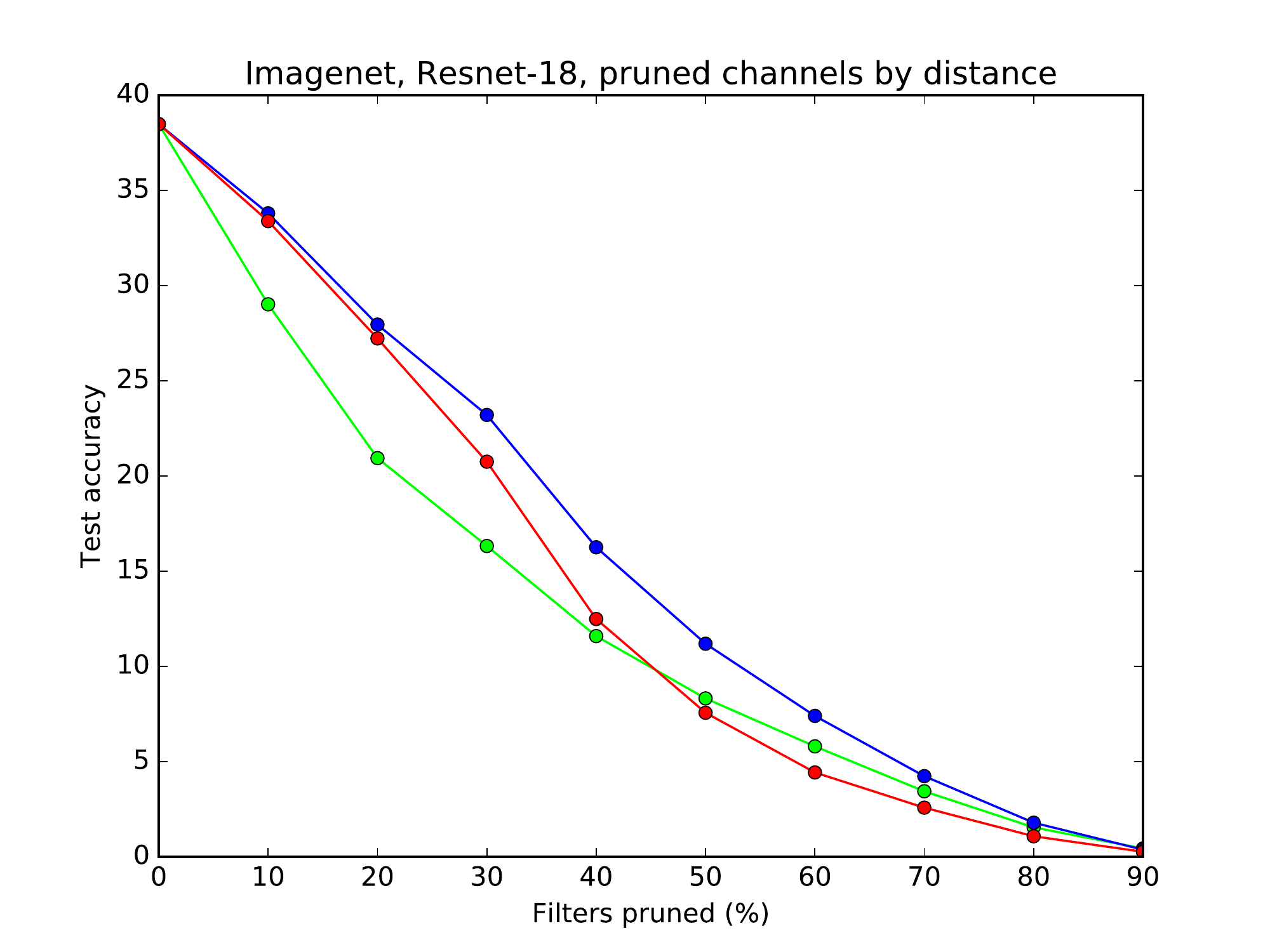}&
    \includegraphics{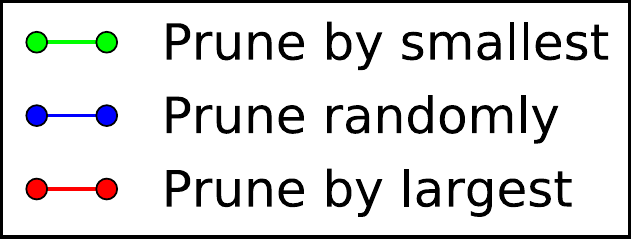} \\
    \includegraphics{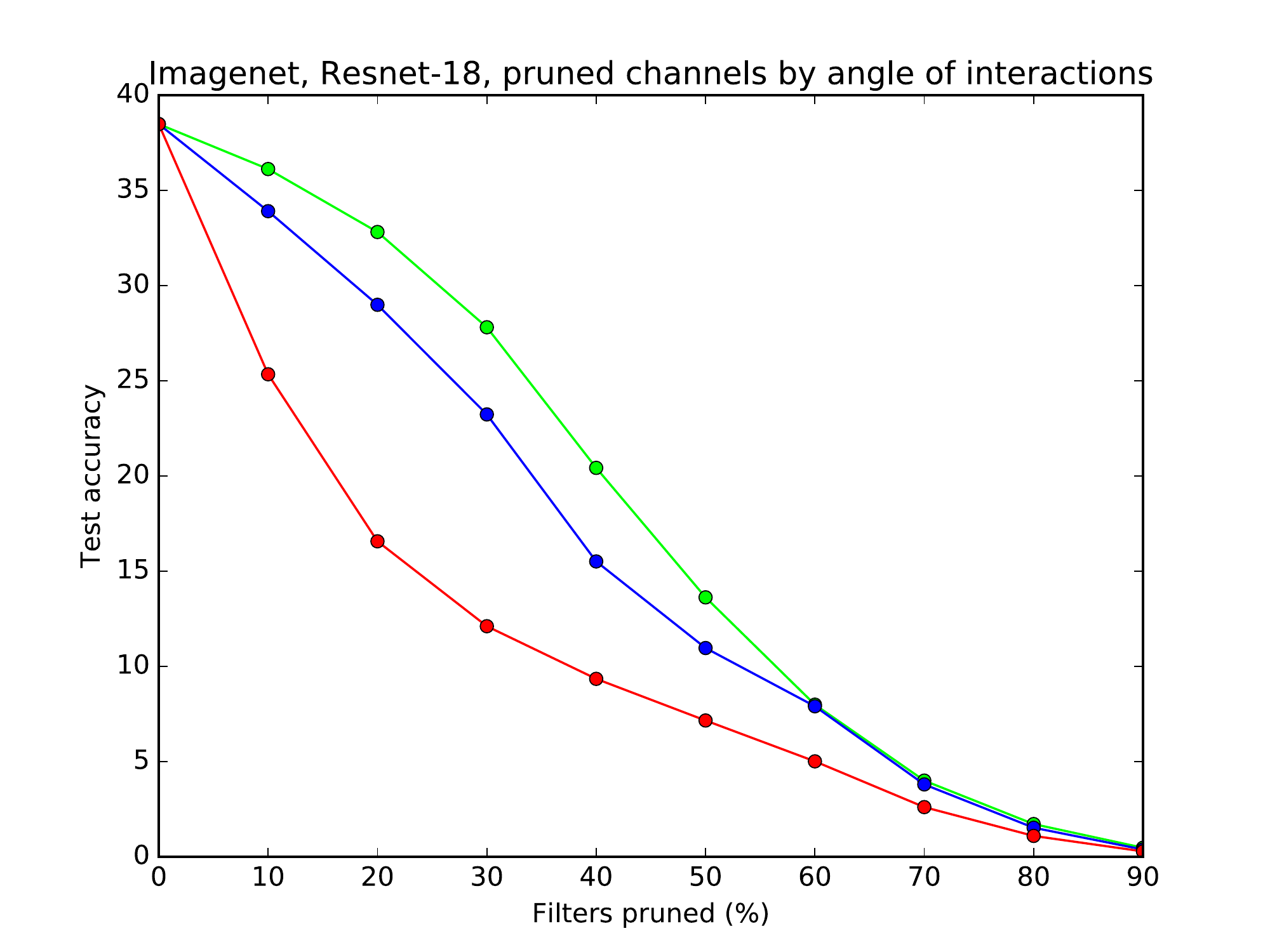}&
    \includegraphics{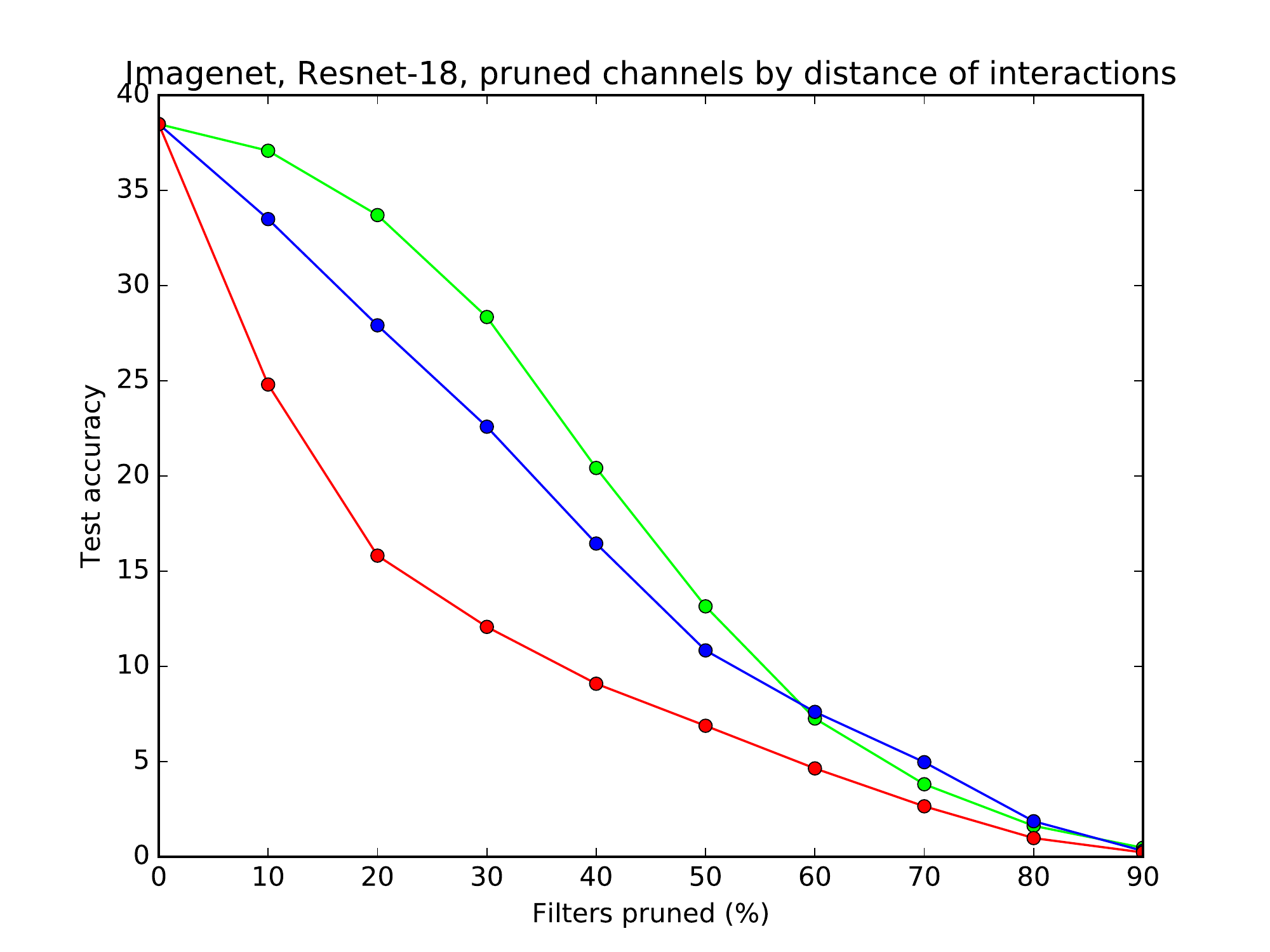}
    \end{tabular}}
    \caption{Pruning complete filters in a BNN ResNet-18 trained on Imagenet. Accuracy during pruning averaged across all the layers using: (a) Angle, (b) Euclidean distance, (c) Angle of the interactions and (d) Euclidean distance of the interactions.}
    \label{fig:pruning_channels}
\end{figure}

Pruning kernels using the proposed metrics proved to be an effective technique, however to avoid working with sparse matrices, it is desirable to prune whole filters at once, hence we extend the method to filter-level. Unlike kernel-level pruning, filter-level pruning removes the corresponding output features maps produced by the pruned filters in the $l$-th layer as well as the related filters in ($l+1$)-th layer. A straightforward way to rank the $\bF_k$ filter is to compute filter-wise distances, either by vectorizing the whole filter and computing the distance or by averaging the kernel-wise distances across the output channel (shaded region in the left side of Figure \ref{fig:convolutional_interactions}). Unexpectedly, this strategy led to poor filter selection as can be observed in Figure~\ref{fig:pruning_channels}. Alternatively, we analyze the interaction of the $l$-th layer with the ($l+1$)-th layer. In specific, we rank the $\bF_k$ filter in the $l$-th layer by averaging the kernel-wise distances across the input channels in the ($l+1$)-th layer (shaded region in the right side of Figure~\ref{fig:convolutional_interactions}). This method led to a good filter selection criteria. We conjecture that if the feature map $\bO_k$ produced by the $k$-th filter is not a meaningful representation, the related kernels in the next layer will not converge.

\subsection{Determining per-layer pruning ratio}  \label{sec:ratio} \label{subsec:bayesian_optimization}
In order to automatically determine a near optimal pruning ratio for each layer, we leverage Bayesian optimization as proposed by Snoek~\etal\cite{snoek2012practical}, for more on GPs we refer the reader to Rasmussen and Williams \cite{rasmussen2005gaussian}. 

Let the objective function for the $l$-th layer be defined as
\begin{equation}
y(x) = \cL_c(x) + \alpha_1\cL_{params}(x) + \alpha_2\cL_{size}(x),
\label{eq:loss_function}
\end{equation}
where $\cL_c$, $\cL_{params}$ and $\cL_{size}$ denote the loss terms incurred by the classification error, number of parameters and network size given all in percentages, and $x$ is the pruning ratio to be estimated. The second and third term are only apparently redundant as will be explained at the end of the section.

Bayesian optimization allows to optimize non-convex, expensive to evaluate, black-box functions by assuming the optimization function was sampled from a Gaussian Process (GP). In this process, multiple pruning ratios are tested for each layer, the evaluation is done in a small subset of the data. In order to pick the pruning ratios to test, an acquisition function can be computed in closed form based on the previous observations.

GPs define distributions over functions of the form $GP:\raisebox{2pt}{$\chi$} \rightarrow \mathbb{R}$. Any given point $x \in \raisebox{2pt}{$\chi$}$ corresponds to a single Gaussian random variable defined by it's mean and covariance functions:
\begin{equation}
m_x = \mathbb{E}\{ x \}.
\label{eq:mean}
\end{equation}
\begin{equation}
c_{x_1x_2} = \mathbb{E}\{ (x_1-m_{x_1})(x_2-m_{x_2}) \}.
\label{eq:covariance}
\end{equation}

The Gaussian Process Upper Confidence Bound (UCB) will be our choice of acquisition function. Alternative acquisition functions include Probability of Improvement (PI) and Expected Improvement (EI) \cite{snoek2012practical}. We define a set of $N$ pruning ratio observations $X = \{x_n\}^N_{n=1}$, with their respective total loss evaluations $Y = \{y_n\}^N_{n=1}$ as defined by Eq. (\ref{eq:loss_function}). Finally, the UCB depends on the predictive mean and predictive variance functions:
\begin{equation}
\mu(\hat{x}) = m_{\hat{x}} + c_{\hat{x}X}C^{-1}_N(Y-m_X),
\label{eq:predictive_mean}
\end{equation}
\begin{equation}
\sigma^2(\hat{x}) = c_{\hat{x}\hat{x}} - c_{\hat{x}X}C^{-1}_Nc_{X\hat{x}},
\label{eq:predictive_variance}
\end{equation}
where $C^{-1}_N$ denotes the inverse covariance matrix of the observations or \textit{information matrix}, and $\hat{x}$ denotes the unexplored candidates. Finally, the UCB has the form
\begin{equation}
UCB(\hat{x}) = \mu(\hat{x})-\kappa\sigma(\hat{x}),
\label{eq:UCB}
\end{equation}
where the parameter $\kappa$ can be tuned to favor exploration over exploitation. Given \raisebox{2pt}{$\chi$} is a bounded region, the acquisition function determines the next observation to be made
\begin{equation}
\hat{x}_{next}=\argmax_{\hat{x}}UCB(\hat{x}).
\end{equation}
Given a limited budget of exploration steps, the $\hat{x}$ that provides the minimum loss according to Eq. (\ref{eq:loss_function}), will be used as the pruning ratio for the $l$-th layer.

\textbf{Compound loss function in Eq.~(\ref{eq:loss_function}).} Conventionally in QNNs, the first and last layers remain with full precision in order to avoid excessive loss of information \cite{rastegari2016xnor}. In our experiments we noticed the second last layer is highly sensitive to pruning, particularly in Xnor-Net and DoReFa quantization schemes, as shown in Table \ref{tab:pruning_layerwise}, therefore, the third loss term is to encourage more aggressive pruning in the full precision weights (\textit{i.e.}, in the first and second last layers). Removing the second term will cause the classification loss to overwhelm the pruning loss in the intermediate layers and no pruning will occur. Furthermore, the opposite situation will occur in the first and second last layers, where the pruning loss will overwhelm the classification loss degrading the accuracy to an unrecoverable state. In practice we found $\alpha_2 = \alpha_1/4$ to be a good setting.

\textbf{Pruning full-precision layers.}
In order to prune the first layer, we use the angle (or distance accordingly) of the interactions in the second layer. Similarly, the second last layer will always use the sub-optimal metric based on the current layer.

\begin{table*}[t]
	\small
	\begin{center}
	\caption{Layerwise pruning statistics for Xnor-Net ResNet-18 on Imagenet. Original accuracy: 52.23, Pruned accuracy: 49.48 ($\star$ Pruning ratio for downsample residual connection will be the same as the contiguous layer).}
		\begin{tabular}{l c c c l c c c}
			\hline
			Layer&Parameters&Size(MB)&Ratio(\%)&Layer&Parameters&Size(MB)&Ratio(\%)\\
			\hline
			conv1 & 9K & 0.035 & 0 & residual2 & 32K & 0.003 & $\star$ \\
			conv2 & 36K & 0.004 & 0 & conv11 & 589K & 0.070 & 1.1 \\
			conv3 & 36K & 0.004 & 0 & conv12 & 589K & 0.070 & 25.53 \\
			conv4 & 36K & 0.004 & 0 & conv13 & 589K & 0.070 & 9.13 \\
			conv5 & 36K & 0.004 & 0 & conv14 & 1179K & 0.140 & 26.29 \\
			conv6 & 73K & 0.008 & 0 & residual3 & 131K & 0.015 & $\star$ \\
			residual1 & 8K & 0.001 & $\star$ & conv15 & 2359K & 0.281 & 1.8 \\
			conv7 & 147K & 0.017 & 0 & conv16 & 2359K & 0.281 & 40.0 \\
			conv8 & 147K & 0.017 & 1.41 & conv17 & 2359K & 0.281 & 0.5 \\
			conv9 & 147K & 0.017 & 1.46 & fc & 513K & 1.956 & 0 \\
			conv10 & 294K & 0.035 & 0 &  &  &  &  \\
			\hline
		\end{tabular}
		\label{tab:pruning_layerwise}
	\end{center}
	\vspace{-1.5em}
\end{table*}

\section{Experiments}
For our experiments we tested 2 different achitectures, VGG \cite{simonyan2014very} and ResNet \cite{he2016deep}, for the task of image classification on the CIFAR-10 \cite{krizhevsky2014cifar} and ImageNet (ILSVRC-2012) \cite{russakovsky2015imagenet} datasets with 4 different quantization schemes, BinaryConnect (full-precision activations and binary weights) \cite{courbariaux2015binaryconnect}, Binarized Neural Networks (binary activations and binary weights) \cite{hubara2016binarized}, Xnor-Net (binary scaled activations and binary scaled weights) \cite{rastegari2016xnor} and DoReFa-Net (2-bit activations and 2-bit weights for our particular experiments) \cite{zhou2016dorefa}. 

Our QNNs implementation was done using the Pytorch framework and the Bayesian optimization using \cite{snoek2012practical} public Bayesian optimization package. To implement our algorithm every convolutional, fully-connected (FC) and Batch Normalization (BN) \cite{ioffe2015batch} layer is assigned two boolean masks, one is used to index the subset of neurons to compute and a second one to index the subset of inputs to use. Residual connections were pruned accordingly where needed. To pre-train our baselines, we trained from scratch for CIFAR-10. For ImageNet we use a full-precision pre-trained model and fine-tune for every specific quantization scheme. We used the standard pre-processing and augmentation as reported by \cite{he2016deep}. For training with the pre-trained full-precision model, we use an initial learning rate of 0.005 and decrease it by a factor of 10 every 10 epochs during 60 epochs.

For the pruning strategy, we tested a layer-wise bottom-up and top-down approach with no noticeable difference. Every time a layer is pruned a non-trivial amount (for our experiments we use above 5\%), the model is fine tuned for 5 epochs to allow the weights to adjust to the new architecture and recover from the accuracy drop. It is worth noting that most of the layers are not pruned, commonly only the later layers will be pruned as can be verified in Table~\ref{tab:pruning_layerwise}, which is reasonable since the input channels increase with the depth of the network, therefore the reward for pruning a deeper layer will be larger. It also suggest that filters are more redundant as the depth increases. For fine-tuning we use the same learning rate schedule used during training but decreasing every epoch. Finally, at the end of the pruning process, additional fine-tuning for another 10 epochs is performed and the best validation value is reported. No guidance loss was used in order to isolate the effect of our method.

In our experiments we report the original accuracy, the accuracy after pruning and fine-tuning, the pruned ratio (the pruned ratio refers to model size in bits, not the number of parameters) and average speedup computed from 10 forward passes. To compute the size of the model we used 1 bit per connection for BNNs and 2 bits for DoReFa-Net, 32 bits for biases and other full-precision parameters (\textit{e.g.}, Xnor-Net scalings, BN and PRelu parameters).

\textbf{Fully binary network.}
As mentioned in Sec.~\ref{subsec:bayesian_optimization}, it is common practice in quantized networks to keep full-precision weights in the first and last layers. By quantizing only the weights in the first layer, it will incur into a small accuracy degradation. In the binary case, hence, the dot product can be efficiently computed exclusively with additions. Similarly, in our experiments we quantized the last layer and placed a BN layer in-between this and the softmax layer. Once again, we noticed it did not incur into a large accuracy degradation. By quantizing both the input and last layers, an extremely light and efficient network can be attained with a tolerable accuracy degradation, and it can be further pruned with out proposed method. We refer to this configuration as fully Binarized Neural Network.

\textbf{Details on Bayesian optimization.} For our layer-wise pruning ratio optimization, we use a bounded region of [0, 40]\% and hyperparameter $\kappa = 2.5$. The process performs 5 observations and 5 exploration steps.

\subsection{Experiments on CIFAR-10}
The CIFAR-10 dataset consists of 50,000 training images plus 10,000 images for validation of size $32 \times 32$ distributed across 10 categories for classification.
VGG and ResNet architectures were designed for the ImageNet dataset, where commonly the images are cropped to size $224 \times 224$, which is considerably larger than CIFAR-10. Therefore, in order to make the pruning task challenging, we utilized compact versions of both. Our VGG-11 network is identical to the original configuration in \cite{simonyan2014very} with the exception that our implementation only has one FC layer. Similarly, our ResNet-14 implementation is a compact version of the original ResNet-18 \cite{he2016deep} where the last stage (a set of 4 \textit{Basicblock} layers) was omitted. The remaining layers were slightly widen by an inflation ratio of 1.25x. For this section we used the pruning loss parameter $\alpha_1=1$, except for fully BNN where we used $\alpha_1=2$.
Our resulting VGG-11 network is still highly overparameterized for the task at hand, thus it is more prunable than ResNet-14. As shown in Table~\ref{tab:vgg11_CIFAR-10} and Table~\ref{tab:ResNet14_CIFAR-10}, a BNN VGG-11 can be further compressed 43\% with no loss in accuracy, however a ResNet-14 BNN can achieve higher accuracy with approximately one third less the size.

\begin{table*}[t]
	\small
	\begin{center}
	\caption{Top-1 accuracy for VGG-11 on CIFAR-10.}
	\scalebox{0.9}{
		\begin{tabular}{l c c c c c}
			\hline
			&Original Acc.(\%) &Retrain Acc.(\%)&Pruned Ratio(\%) &Memory(MB) &Speedup \\
			\hline
			BinaryConnect & 87.60 & 86.53 & 34.44 & 0.75 & 1.4x \\
			BNN & 82.07 & 82.31 & 43.82 & 0.64 & 1.4x \\
			BNN (fully) & 84.02 & 83.09 & 48.59 & 0.58 & 1.5x \\
			Xnor-Net & 78.80 & 74.01 & 26.41 & 0.88 & 1.1x \\
			DoReFa-Net & 87.27 & 86.30 & 53.73 & 0.94 & 1.4x \\
			\hline
		\end{tabular}}
		\label{tab:vgg11_CIFAR-10}
     \end{center}
     \vspace{-1.0em}
\end{table*}

\begin{table*}[t]
	\small
	\begin{center}
	\caption{Top-1 accuracy for ResNet-14 on CIFAR-10.}
	\scalebox{0.9}{
		\begin{tabular}{l c c c c c}
			\hline
			&Original Acc.(\%) &Retrain Acc.(\%)&Pruned Ratio(\%) &Memory(MB) &Speedup \\
			\hline
			BinaryConnect & 92.73 & 91.76 & 21.79 & 0.43 & 1.4x \\
			BNN & 89.47 & 88.76 & 26.79 & 0.40 & 1.1x \\
			BNN (fully) & 88.26 & 87.58 & 31.20 & 0.37 & 1.4x \\
			Xnor-Net & 86.59 & 83.90 & 21.61 & 0.43 & 1.3x \\
			DoReFa-Net & 90.58 & 89.84 & 31.34 & 0.73 & 1.2x \\
			\hline
		\end{tabular}}
		\vspace{-2em}
		\label{tab:ResNet14_CIFAR-10}
	\end{center}
\end{table*}

In order to compare with different metrics, we performed tests with ResNet-14 using three different quantizations: BNN, fully BNN and DoReFa. As reported by \cite{hubara2016quantized} and \cite{leroux2019training}, our BNN and Xnor-Net implementations achieved comparable results, frequently BNN outperforming Xnor-Net, therefore we will use BNN quantization for our comparison of the different metrics due to its simplicity and efficiency. As can be observed in Table \ref{tab:metrics_comparison} for a small dataset like CIFAR, all the distance metrics performed reasonable well.

\begin{table*}[t]
	\small
	\centering
		\caption{Top-1 accuracy for ResNet-14 on CIFAR-10.}
	\scalebox{0.9}{
		\begin{tabular}{l c c c c c}
			\specialrule{.2em}{.1em}{.1em}
			\multicolumn{6}{c}{\bfseries Pruning filters by angle of the filter} \\
			\hline
			&Original Acc.(\%) &Retrain Acc.(\%)&Pruned Ratio(\%) &Memory(MB) &Speedup \\
			\hline
			BNN & 89.47 & 87.71 & 28.17 & 0.39 & 1.4x \\
			BNN (full) & 88.26 & 87.89 & 29.37 & 0.38 & 1.4x \\   % alpha = 9
			DoReFa-Net & 90.58 & 89.23 & 34.48 & 0.69 & 1.3x \\
			\hline
			\specialrule{.2em}{.1em}{.1em}
			\multicolumn{6}{c}{\bfseries Pruning filters by euclidean distance of the filter} \\
			\hline
			&Original Acc.(\%) &Retrain Acc.(\%)&Pruned Ratio(\%) &Memory(MB) &Speedup \\
			\hline
			BNN & 89.47 & 87.98 & 29.02 & 0.39 & 1.4x \\
			BNN (full) & 88.26 & 87.70 & 37.17 & 0.33 & 1.4x \\
			DoReFa-Net & 90.58 & 89.40 & 23.38 & 0.81 & 1.1x \\
			\hline
			\specialrule{.2em}{.1em}{.1em}
			\multicolumn{6}{c}{\bfseries Pruning filters by angle of the interactions} \\
			\hline
			&Original Acc.(\%) &Retrain Acc.(\%)&Pruned Ratio(\%) &Memory(MB) &Speedup \\
			\hline
			BNN & 89.47 & 88.76 & 26.79 & 0.40 & 1.1x \\ % alpha = 8
			BNN (full) & 88.26 & 87.58 & 31.20 & 0.37 & 1.4x \\ % alpha = 8
			DoReFa-Net & 90.58 & 89.84 & 31.34 & 0.73 & 1.2x \\ % alpha = 5
			\hline
			\specialrule{.2em}{.1em}{.1em}
			\multicolumn{6}{c}{\bfseries Pruning filters by euclidean distance of the interactions} \\
			\hline
			&Original Acc.(\%) &Retrain Acc.(\%)&Pruned Ratio(\%) &Memory(MB) &Speedup \\
			\hline
			BNN & 89.47 & 88.50 & 31.70 & 0.38 & 1.4x \\
			BNN (full) & 88.26 & 87.56 & 40.51 & 0.32 & 1.5x \\
			DoReFa-Net & 90.58 & 89.15 & 22.10 & 0.83 & 1.1x \\
			\hline
		\end{tabular}}
		\label{tab:metrics_comparison}
	\vspace{-1.0em}
\end{table*}

\subsection{Experiments on ImageNet}
The ImageNet dataset \cite{russakovsky2015imagenet} contains 1.28M labelled images for training and 50K for validation for the task of object classification, spread across 1,000 classes.
For the experiments in this section, we used a standard ResNet-18 architecture with no inflation. PRelu layers were placed after the quantized convolutional layers which provided a small increase in accuracy. We pre-train a model using the described procedure and afterwards we proceed to the pruning stage. For this section we used the pruning loss parameter $\alpha_1=8$.

Table \ref{tab:overall_results} shows the results of our pruning strategy using the angle of the interactions for ranking. We did not find significant differences when using the angles and the euclidean distance of the interactions, therefore the decision of using either metric is arbitrary.
We compare our method only to \cite{xu2018main} on pruning Xnor-Net given that there is no baseline for the rest of the quantization schemes. \cite{xu2018main} achieved 21.40\% parameter pruning corresponding to 8.39\% of the total model size with a top-1 accuracy of 50.13\%. Our method achieved 25.5\% parameter pruning equivalent to 10.63\% of the total size and top-1 accuracy of 49.48\%. It is worth noting that our original network is lighter, therefore, although our pruning ratio is only moderately higher, our pruned network is considerably lighter (3.85 MB - 3.01 MB). 

Finally, we additionally pruned a BNN to achieve an extremely-light network and DoReFa-Net to demonstrate the scalability of our method to generic quantization. In our experiments we noticed DoReFa-Net is slightly more sensitive to pruning than the other quantization schemes.

\begin{table*}[t]
	\small
	\begin{center}
		\caption{Overall results for ResNet-18 on ImageNet using the angle of the interactions.}
	    \scalebox{1.0}
		{\begin{tabular}{l c c c c c}
			\hline
			&Orig. Acc.(\%)&Ret. Acc.(\%)&P. Ratio(\%)&Memory(MB)&Speedup \\
			\hline
			BinaryConnect & 61.23 & 58.38 & 20.66 & 2.66 & 1.4x \\ %alpha = 8
			%BinaryConnect & 61.23 & 58.76 & 18.45 & 2.73 & 1.4x \\ %alpha = 7
			BNN & 51.92 & 50.10 & 16.69 & 2.73 & 1.4x \\ % alpha = 8
			BNN (fully) & 35.80 & 31.58 & 10.27 & 1.29 & 0x \\
			Xnor-Net & 52.23 & 49.48 & 10.63 & 3.01 & 1.2x \\ % alpha = 5
			Xnor-Net pruned by \cite{xu2018main} & 49.98 & 50.13 & 8.39 & 3.85 & 1.09x \\
			DoReFa-Net & 60.03 & 59.30 & 6.46 & 4.36 & 1.4x \\ % alpha = 5
			\hline
		\end{tabular}}
		\label{tab:overall_results}
	\end{center}
	\vspace{-1.0em}
\end{table*}

\section{Conclusion}
In this paper, we have proposed to prune quantized neural networks to yield extremely energy-efficient yet accurate models for embedded devices. In specific, we have used the shadow full-precision model along with its quantized version in order to accurately approximate full-precision convolutions. To measure the importance of each filter, we have proposed two metrics, including cosine distance and euclidean distance. During pruning, we leverage the interactions between two consecutive layers. We have conducted extensive experiments on CIFAR-10 and ImageNet classification task and observed that conventional quantized architectures can be further pruned with low degradation.

\section{Acknowledgement}
This work was supported by the Australian Research Council through the ARC Centre of Excellence for Robotic Vision (project number CE1401000016).

{\small
\bibliographystyle{ieee_fullname}
\bibliography{egbib}

\begin{thebibliography}{10}\itemsep=-1pt

\bibitem{anderson2017high}
Alexander~G Anderson and Cory~P Berg.
\newblock The high-dimensional geometry of binary neural networks.
\newblock {\em arXiv preprint arXiv:1705.07199}, 2017.

\bibitem{ashok2018n2n}
Anubhav Ashok, Nicholas Rhinehart, Fares Beainy, and Kris~M Kitani.
\newblock N2n learning: Network to network compression via policy gradient
  reinforcement learning.
\newblock In {\em Proc. Int. Conf. Learn. Repren.}, 2018.

\bibitem{bai2019proxquant}
Yu Bai, Yu-Xiang Wang, and Edo Liberty.
\newblock Proxquant: Quantized neural networks via proximal operators.
\newblock In {\em Proc. Int. Conf. Learn. Repren.}, 2019.

\bibitem{Cai_2017_CVPR}
Zhaowei Cai, Xiaodong He, Jian Sun, and Nuno Vasconcelos.
\newblock Deep learning with low precision by half-wave gaussian quantization.
\newblock In {\em Proc. IEEE Conf. Comp. Vis. Patt. Recogn.}, pages 5918--5926,
  2017.

\bibitem{courbariaux2015binaryconnect}
Matthieu Courbariaux, Yoshua Bengio, and Jean-Pierre David.
\newblock Binaryconnect: Training deep neural networks with binary weights
  during propagations.
\newblock In {\em Proc. Adv. Neural Inf. Process. Syst.}, pages 3123--3131,
  2015.

\bibitem{guo2017network}
Yiwen Guo, Anbang Yao, Hao Zhao, and Yurong Chen.
\newblock Network sketching: Exploiting binary structure in deep cnns.
\newblock In {\em Proc. IEEE Conf. Comp. Vis. Patt. Recogn.}, pages 5955--5963,
  2017.

\bibitem{he2016deep}
Kaiming He, Xiangyu Zhang, Shaoqing Ren, and Jian Sun.
\newblock Deep residual learning for image recognition.
\newblock In {\em Proc. IEEE Conf. Comp. Vis. Patt. Recogn.}, pages 770--778,
  2016.

\bibitem{he2018amc}
Yihui He, Ji Lin, Zhijian Liu, Hanrui Wang, Li-Jia Li, and Song Han.
\newblock Amc: Automl for model compression and acceleration on mobile devices.
\newblock In {\em Proc. Eur. Conf. Comp. Vis.}, pages 784--800, 2018.

\bibitem{he2017channel}
Yihui He, Xiangyu Zhang, and Jian Sun.
\newblock Channel pruning for accelerating very deep neural networks.
\newblock In {\em Proc. IEEE Int. Conf. Comp. Vis.}, volume~2, page~6, 2017.

\bibitem{hubara2016binarized}
Itay Hubara, Matthieu Courbariaux, Daniel Soudry, Ran El-Yaniv, and Yoshua
  Bengio.
\newblock Binarized neural networks.
\newblock In {\em Proc. Adv. Neural Inf. Process. Syst.}, pages 4107--4115,
  2016.

\bibitem{hubara2016quantized}
Itay Hubara, Matthieu Courbariaux, Daniel Soudry, Ran El-Yaniv, and Yoshua
  Bengio.
\newblock Quantized neural networks: Training neural networks with low
  precision weights and activations.
\newblock {\em arXiv preprint arXiv:1609.07061}, 2016.

\bibitem{ioffe2015batch}
Sergey Ioffe and Christian Szegedy.
\newblock Batch normalization: Accelerating deep network training by reducing
  internal covariate shift.
\newblock In {\em Proc. Int. Conf. Mach. Learn.}, pages 448--456, 2015.

\bibitem{jung2018joint}
Sangil Jung, Changyong Son, Seohyung Lee, Jinwoo Son, Youngjun Kwak, Jae-Joon
  Han, and Changkyu Choi.
\newblock Joint training of low-precision neural network with quantization
  interval parameters.
\newblock {\em arXiv preprint arXiv:1808.05779}, 2018.

\bibitem{krizhevsky2014cifar}
Alex Krizhevsky, Vinod Nair, and Geoffrey Hinton.
\newblock The cifar-10 dataset.
\newblock {\em online: http://www. cs. toronto. edu/kriz/cifar. html}, 55,
  2014.

\bibitem{lee2019snip}
Namhoon Lee, Thalaiyasingam Ajanthan, and Philip~HS Torr.
\newblock Snip: Single-shot network pruning based on connection sensitivity.
\newblock In {\em Proc. Int. Conf. Learn. Repren.}, 2019.

\bibitem{leroux2019training}
Sam Leroux, Bert Vankeirsbilck, Tim Verbelen, Pieter Simoens, and Bart Dhoedt.
\newblock Training binary neural networks with knowledge transfer.
\newblock {\em Neurocomputing}, 2019.

\bibitem{li2016pruning}
Hao Li, Asim Kadav, Igor Durdanovic, Hanan Samet, and Hans~Peter Graf.
\newblock Pruning filters for efficient convnets.
\newblock In {\em Proc. Int. Conf. Learn. Repren.}, 2017.

\bibitem{li2017performance}
Zefan Li, Bingbing Ni, Wenjun Zhang, Xiaokang Yang, and Wen Gao.
\newblock Performance guaranteed network acceleration via high-order residual
  quantization.
\newblock In {\em Proc. IEEE Int. Conf. Comp. Vis.}, pages 2584--2592, 2017.

\bibitem{lin2017refinenet}
Guosheng Lin, Anton Milan, Chunhua Shen, and Ian Reid.
\newblock Refinenet: Multi-path refinement networks for high-resolution
  semantic segmentation.
\newblock In {\em Proceedings of the IEEE conference on computer vision and
  pattern recognition}, pages 1925--1934, 2017.

\bibitem{lin2017towards}
Xiaofan Lin, Cong Zhao, and Wei Pan.
\newblock Towards accurate binary convolutional neural network.
\newblock In {\em Proc. Adv. Neural Inf. Process. Syst.}, pages 344--352, 2017.

\bibitem{liu2017learning}
Zhuang Liu, Jianguo Li, Zhiqiang Shen, Gao Huang, Shoumeng Yan, and Changshui
  Zhang.
\newblock Learning efficient convolutional networks through network slimming.
\newblock In {\em Proc. IEEE Int. Conf. Comp. Vis.}, pages 2755--2763. IEEE,
  2017.

\bibitem{long2015fully}
Jonathan Long, Evan Shelhamer, and Trevor Darrell.
\newblock Fully convolutional networks for semantic segmentation.
\newblock In {\em Proc. IEEE Conf. Comp. Vis. Patt. Recogn.}, pages 3431--3440,
  2015.

\bibitem{louizos2019relaxed}
Christos Louizos, Matthias Reisser, Tijmen Blankevoort, Efstratios Gavves, and
  Max Welling.
\newblock Relaxed quantization for discretized neural networks.
\newblock In {\em Proc. Int. Conf. Learn. Repren.}, 2019.

\bibitem{louizos2017learning}
Christos Louizos, Max Welling, and Diederik~P Kingma.
\newblock Learning sparse neural networks through $ l\_0 $ regularization.
\newblock In {\em Proc. Int. Conf. Learn. Repren.}, 2018.

\bibitem{luo2017thinet}
Jian-Hao Luo, Jianxin Wu, and Weiyao Lin.
\newblock Thinet: A filter level pruning method for deep neural network
  compression.
\newblock In {\em Proc. IEEE Int. Conf. Comp. Vis.}, pages 5058--5066, 2017.

\bibitem{rasmussen2005gaussian}
Carl~Edward Rasmussen and Christopher K.~I. Williams.
\newblock {\em Gaussian Processes for Machine Learning (Adaptive Computation
  and Machine Learning)}.
\newblock The MIT Press, 2005.

\bibitem{rastegari2016xnor}
Mohammad Rastegari, Vicente Ordonez, Joseph Redmon, and Ali Farhadi.
\newblock Xnor-net: Imagenet classification using binary convolutional neural
  networks.
\newblock In {\em Proc. Eur. Conf. Comp. Vis.}, pages 525--542, 2016.

\bibitem{russakovsky2015imagenet}
Olga Russakovsky, Jia Deng, Hao Su, Jonathan Krause, Sanjeev Satheesh, Sean Ma,
  Zhiheng Huang, Andrej Karpathy, Aditya Khosla, Michael Bernstein, et~al.
\newblock Imagenet large scale visual recognition challenge.
\newblock {\em Int. J. Comp. Vis.}, 115(3):211--252, 2015.

\bibitem{simonyan2014very}
Karen Simonyan and Andrew Zisserman.
\newblock {Very deep convolutional networks for large-scale image recognition}.
\newblock In {\em Proc. Int. Conf. Learn. Repren.}, 2015.

\bibitem{snoek2012practical}
Jasper Snoek, Hugo Larochelle, and Ryan~P Adams.
\newblock Practical bayesian optimization of machine learning algorithms.
\newblock In {\em Proc. Adv. Neural Inf. Process. Syst.}, pages 2951--2959,
  2012.

\bibitem{sutskever2014sequence}
Ilya Sutskever, Oriol Vinyals, and Quoc~V Le.
\newblock Sequence to sequence learning with neural networks.
\newblock In {\em Advances in neural information processing systems}, pages
  3104--3112, 2014.

\bibitem{tang2017train}
Wei Tang, Gang Hua, and Liang Wang.
\newblock How to train a compact binary neural network with high accuracy?
\newblock In {\em Proc. AAAI Conf. on Arti. Intel.}, pages 2625--2631, 2017.

\bibitem{tung2018clip}
Frederick Tung and Greg Mori.
\newblock Clip-q: Deep network compression learning by in-parallel
  pruning-quantization.
\newblock In {\em Proceedings of the IEEE Conference on Computer Vision and
  Pattern Recognition}, pages 7873--7882, 2018.

\bibitem{vaswani2017attention}
Ashish Vaswani, Noam Shazeer, Niki Parmar, Jakob Uszkoreit, Llion Jones,
  Aidan~N Gomez, {\L}ukasz Kaiser, and Illia Polosukhin.
\newblock Attention is all you need.
\newblock In {\em Advances in neural information processing systems}, pages
  5998--6008, 2017.

\bibitem{wang2018skipnet}
Xin Wang, Fisher Yu, Zi-Yi Dou, Trevor Darrell, and Joseph~E Gonzalez.
\newblock Skipnet: Learning dynamic routing in convolutional networks.
\newblock In {\em Proc. Eur. Conf. Comp. Vis.}, pages 409--424, 2018.

\bibitem{wu2018blockdrop}
Zuxuan Wu, Tushar Nagarajan, Abhishek Kumar, Steven Rennie, Larry~S Davis,
  Kristen Grauman, and Rogerio Feris.
\newblock Blockdrop: Dynamic inference paths in residual networks.
\newblock In {\em Proc. IEEE Conf. Comp. Vis. Patt. Recogn.}, pages 8817--8826,
  2018.

\bibitem{xu2018main}
Yinghao Xu, Xin Dong, Yudian Li, and Hao Su.
\newblock A main/subsidiary network framework for simplifying binary neural
  network.
\newblock In {\em Proc. IEEE Conf. Comp. Vis. Patt. Recogn.}, 2019.

\bibitem{yu2018nisp}
Ruichi Yu, Ang Li, Chun-Fu Chen, Jui-Hsin Lai, Vlad~I Morariu, Xintong Han,
  Mingfei Gao, Ching-Yung Lin, and Larry~S Davis.
\newblock Nisp: Pruning networks using neuron importance score propagation.
\newblock In {\em Proc. IEEE Conf. Comp. Vis. Patt. Recogn.}, pages 9194--9203,
  2018.

\bibitem{zhang2018lq}
Dongqing Zhang, Jiaolong Yang, Dongqiangzi Ye, and Gang Hua.
\newblock Lq-nets: Learned quantization for highly accurate and compact deep
  neural networks.
\newblock In {\em Proc. Eur. Conf. Comp. Vis.}, 2018.

\bibitem{zhou2016dorefa}
Shuchang Zhou, Yuxin Wu, Zekun Ni, Xinyu Zhou, He Wen, and Yuheng Zou.
\newblock Dorefa-net: Training low bitwidth convolutional neural networks with
  low bitwidth gradients.
\newblock {\em arXiv preprint arXiv:1606.06160}, 2016.

\bibitem{zhuang2018towards}
Bohan Zhuang, Chunhua Shen, Mingkui Tan, Lingqiao Liu, and Ian Reid.
\newblock Towards effective low-bitwidth convolutional neural networks.
\newblock In {\em Proc. IEEE Conf. Comp. Vis. Patt. Recogn.}, 2018.

\bibitem{zhuang2019structured}
Bohan Zhuang, Chunhua Shen, Mingkui Tan, Lingqiao Liu, and Ian Reid.
\newblock Structured binary neural network for accurate image classification
  and semantic segmentation.
\newblock In {\em Proc. IEEE Conf. Comp. Vis. Patt. Recogn.}, 2019.

\bibitem{zhuang2018discrimination}
Zhuangwei Zhuang, Mingkui Tan, Bohan Zhuang, Jing Liu, Yong Guo, Qingyao Wu,
  Junzhou Huang, and Jinhui Zhu.
\newblock Discrimination-aware channel pruning for deep neural networks.
\newblock In {\em Proc. Adv. Neural Inf. Process. Syst.}, 2018.

\end{thebibliography}
}

\end{document}